\documentclass[11pt]{article}
\usepackage{setspace,epsfig,amsmath,amssymb,color,stmaryrd,nicefrac,url}
\usepackage[round]{natbib}
\usepackage{subfig}
\usepackage{algorithmic}
\usepackage[ruled,vlined,linesnumbered]{algorithm2e}
\usepackage{multirow}

\newcommand{\given}{\, \vert \, }
\renewcommand{\vec}[1]{\boldsymbol{#1}}

\newcommand{\stimes}{{\times}}
 
\begin{document}

\title{\bf\LARGE TSK-Streams: Learning TSK Fuzzy Systems\\ on Data Streams}

\author{Ammar Shaker\,${}^{a}$ and Eyke H\"ullermeier\,${}^{b}$\\[3mm]
\small ${}^{a}$\,NEC Laboratories Europe\\[-1mm]
\small Heidelberg, Germany \\[-1mm]
\small ammar.shaker@neclab.eu\\[2mm]
\small ${}^{b}$\,Paderborn University\\[-1mm]
\small Heinz Nixdorf Institute and Department of Computer Science\\[-1mm]
\small Paderborn, Germany\\[-1mm]
\small eyke@upb.de}

\date{}

\maketitle

\begin{abstract}
		
		The problem of adaptive learning from evolving and possibly non-stationary data streams has attracted a lot of interest in machine learning in the recent past, and also stimulated research in related fields, such as computational intelligence and fuzzy systems. In particular, several rule-based methods for the incremental induction of regression models have been proposed. In this paper, we develop a method that combines the strengths of two existing approaches rooted in different learning paradigms. More concretely, our method adopts basic principles of the state-of-the-art learning algorithm AMRules and enriches them by the representational advantages of fuzzy rules. In a comprehensive experimental study, TSK-Streams is shown to be highly competitive in terms of performance.\\[2mm]
		
	\end{abstract}

	\section{Introduction}
	
	In many practical applications of machine learning and predictive modeling, data is produced incrementally in the course of time and observed in the form of a continuous, potentially unbounded stream of observations. Correspondingly, the problem of learning from data streams has recently received increasing attention \citep{Gama:2012:ASLFDSCFT}. Algorithms for learning on streams must be able to process the data in a single pass, which implies an incremental mode of learning, and to adapt to changes of the underlying data-generating process \citep{Domingos:2003:AGFMMDA}. 
	

	
A popular approach for learning on data streams, both for classification and regression, is rule induction, in the fuzzy logic and computational intelligence community also known as ``evolving fuzzy systems''  \citep{Lughofer:2011:EFSMACA}. \cite{mpub360} proposed a method for regression that builds on a very efficient and effective technique for rule induction, which is inspired by the state-of-the-art machine learning algorithm AMRules, and combines it with the strengths of fuzzy modeling. Thus, the method induces a set of fuzzy rules, which, compared to conventional rules with Boolean antecedents, has the advantage of producing smooth regression functions. The method presented in this paper, called TSK-Streams, is a revised and improved variant. The main modifications and novel contributions are as follows.
\begin{itemize}
	\item We give a concise overview of regression learning on data streams as well as a systematic comparison of existing methods with regard to properties such as discretization of features, splitting criteria for rules, etc. This overview helps to better understand the specificities of approaches originating from different research fields, as well as to position our own approach. 
	
	\item We introduce a new strategy for the induction of TSK fuzzy rules and realize it in the form of two concrete variants: variance reduction and error reduction.
While the former is still close to \citep{mpub360}, the variance reduction approach has not been considered for online learning of fuzzy systems so far. Compared with error reduction and other state-of-the-art methods, it leads to models with superior predictive performance. 

	\item In \citep{mpub360}, rule antecedents may contain disjunctions and negations, which makes them difficult to understand and interpret. The representation of TSK rules used in this paper is simpler and more concise. This is achieved by means of an improved technique for splitting fuzzy sets (and extending corresponding rules). 

	
	\item We propose the induction of candidate fuzzy rules using a discretization technique that is based on an extended Binary Search Tree (eBST) structure. 
Compared to the three-layered discretization architecture used by \cite{mpub360}, the use of eBST for constructing candidate fuzzy sets has a number of advantages in the context of online learning. Most notably, it comes with a reduction of complexity from linear to logarithmic (in the number of candidate extensions).

	\item Our empirical evaluation is more extensive and comprises a couple of additional large-scale data sets with up to 100k instances. The evaluation is also extended by including an additional method that has been introduced recently.
\end{itemize}

	
	\section{Related Work}
	
	In the machine learning community, research on supervised learning from data streams has mainly focused on classification problems so far. As one of the first methods, Hoeffding trees \citep{Domingos:2000:MHDS} have been proposed for learning classifiers on high-speed data streams. Since then, the tree-based approach has been developed further, and various modifications and variants can be found in the current literature \citep{Bifet:2009:ALFEDS}. 
	Closely related to tree-based approaches is the induction of decision rules. For example,  the Adaptive Very Fast Decision Rules (AVFDR) method \citep{Kosina:2012:HTCDAVFDR} is an extension of the Very Fast Decision Rules (VFDR) classifier \citep{Gama:2011:LDRDS}, which learns a compact set of rules in an incrementally manner. Most recently, \cite{Bifet:2017:EFDTMEDS} developed an extremely fast version of Hoeffding trees with an implementation that is ready to be used in industrial environments.
	
	Less research has been done on regression for data streams. Notable exceptions include AMRules \citep{Gama:2013:AMRFDS}, which is an extension of AVFDR for handling numeric target values, and FIMTDD \citep{Ikonomovska:2011:LMTFEVDS}, which induces model trees. In contrast to the machine learning community, the fuzzy systems community has put more emphasis on regression than on classification \citep{Angelov:2002:ERM,Angelov:2010:EIS,Lughofer:2011:EFSMACA}. 
	In particular, FLEXFIS \citep{Lughofer:2008:FLEXFISRILAETSFM} is a method for inducing  Takagi-Sugeno-Kang (TSK) rules \citep{Takagi:1985:FISAM} from data streams.

	In the following, we elaborate a bit more on those approaches that are especially relevant for our own method and the experimental study presented later on namely.	
		
	In the Adaptive Model Rules (AMRules) approach, the rule premises are represented in the form of conjunctive combinations of literals on the input variables. Moreover, the rule consequents are specified as linear functions of the variables, which are fitted to the data using least squares regression. Each rule maintains various statistics characterising the part of the instance space covered by that rule. Starting with a single literal, each rule is expanded by new literals step by step, using the Hoeffding bound as a selection criterion. 
	A distinction between unordered rule sets and decision lists is made by \cite{Gama:2013:AMRFDS}. In this paper, the authors propose two prediction and update schemes. 
In the first approach, the rules are sorted in the order in which they have been learned. For prediction, only the first rule that is activated by an example is used. In the second approach, the rules are treated as a set, and their predictions are aggregated in terms of a weighted sum\footnote{While the concrete weight of a rule is not specified in \citep{Gama:2013:AMRFDS}, the implementation suggests that the weight is determined on the basis of the rule's previous errors.}. Moreover, all rules activated by an example are updated. Since a better performance was achieved for the second approach, the authors used that one in their study.
	
	Fast Incremental Model Trees with Drift Detection (FIMTDD) is a method for learning model trees for regression. To determine splits of the model tree, candidate attributes are assessed according to how much they they help to reduce the variance of the target variable. Moreover, a linear function on a corresponding subspace is specified for each leaf of the induced tree, and learning these functions is accomplished using stochastic gradient descent. An ensemble version of FIMTDD (adaptive random forest, ARF-Reg) was proposed by \cite{Gomes:2018:ARFDSR}, using an online version of bagging for creating the ensemble members \citep{Oza:2001:OBB}. 
	
		
	The Fexible Fuzzy Inference Systems (FLEXFIS) approach by \cite{Lughofer:2008:FLEXFISRILAETSFM} is a method for learning fuzzy rules, or, more specifically, Takagi-Sugeno-Kang (TSK) rules, on data streams. This type of rule will be formally introduced in Section 4.1. In contrast to Boolean rules, fuzzy rules are of a gradual nature and can cover an instance to a certain degree, which in turn allows for modulating the influence of a rule on a prediction in a more fine-granular way. In FLEXFIS, the fuzzy support of a rule, i.e., the region it covers in the input space, is determined by (incrementally) clustering the training data and associating each rule with a cluster. Rule consequents are specified in terms of linear functions of the input variables, and the estimation of these functions is successively adapted through recursive weighted least squares (RWLS) \citep{Ljung:1999:SITU}. 
	
	
	The main motivation of our approach is to take advantage of the effectivity and efficiency and algorithmic techniques for rule learning as implemented by methods such as AMRules, and to combine them with the expressiveness of fuzzy rules as used in approaches like FLEXFIS and  eTS+ \citep{Angelov:2010:ETSFSDSeTS+} as well as related formalisms such as fuzzy pattern trees \citep{mpub231}.
	  

	\section{Learning Regression Models on Data Streams}
	\label{Learning Regression Models on Data Streams}
	
	In the following, we categorize and distinguish the learning algorithms discussed above according to several properties. Along the way, we highlight potential advantages of combining different algorithms and their features.

	\subsection{Trees versus rules}
	\label{Trees versus rules}
	Most tree and rule induction methods are based on refining rules in a general-to-specific manner, i.e., they share the property of moving from general to more specific hypotheses. In FIMTDD, for example, leaf nodes are split into more specific leaf nodes. Likewise, in AMRules and TSK-Streams, rules are specialized by adding terms to the premise part. 
		
	Trees can be seen as rule sets with a specific structure. Thus, while a direct transformation from a tree to a set of rules can usually be done in a straight-forward manner, the other direction is not always possible. In AMRules, for example, some of the rules are removed upon detection of a concept change, which makes it impossible to map the current rules to an equivalent tree-model.
	
	FLEXFIS and eTS+ do not follow the aforementioned general-to-specific induction scheme. Instead, they learn and maintain rules in the form of clusters directly in the instance space. In general, these rules cannot be represented in terms of an equivalent tree structure.

	\subsection{Binary versus gradual membership}
	The application of fuzzy logic in decision tree and rule learning leads to two important distinctions from conventional learning.
	First, hard conditions (in rule antecedents) are replaced by soft conditions, so that an example can satisfy a condition to a certain degree. Therefore, in a tree structure, an instance can be propagated to different sibling nodes/leaves simultaneously, perhaps with different weights. Likewise, in a system of rules, it can be covered by multiple rules with different membership degrees.
	
	The second difference is the ability to aggregate the decisions made by different rules in a weighted manner, as done by TSK-Streams, FLEXFIS, and eTS+, instead of merely computing an unweighted average of the outputs of all rules covering an instance. Thus, more weight can be given to the more relevant and less to the less relevant rules.
 
	Likewise, gradual membership allows for more general inference in the case of tree-structures. While decision and model trees restrict tree traversal to a single branch from the root to a leave node, an equivalent fuzzy model tree\footnote{The authors are not aware of any fuzzy model tree induction method for regression on data streams.} would follow several such paths simultaneously, branching an instance at an inner node in a weighted manner depending on how much it agrees with the conditions associated with each branch.

	\subsection{Discretization}
	
	Discretization is usually needed to create a finite number of candidate values for splitting points (thresholds) in the case of continuous features; these splitting points are then validated using a splitting criterion to decide how a tree/rule should be extended.
	
	Both AMRules and FIMTDD apply a supervised discretization technique that is tailored to each rule and leaf node; this is achieved by considering the target values of all instances that reached a given leaf node or are covered by a rule. 
	 
	TSK-Streams, as we will explain later, applies a supervised discretization technique for the creation of fuzzy sets that are evaluated for future extensions.
	
	\subsection{Splitting criteria}
 	
	As already said, refining a model normally means extending a rule with additional conditions, thereby splitting it into two more specific rules or shrinking the region covered by that rule. A splitting criterion is used to find the presumably best among the (typically large) set of candidate splits. To quantify the usefulness of a split, different measures are conceivable.     
	
	A splitting criterion employed by many method, including AMRules, is \emph{variance reduction}: For the rule $R$, the instances $N$ covered by that rule are split into two groups $N_1$ and $N_2$ based on an attribute $x_j$ and a threshold $v$, i.e., 
	\begin{align*}
	N_1 & =\{(\vec{x},y)\in N \, | \, x_{j}\leq v \} \, , \\
	N_2 & =\{(\vec{x},y)\in N \, | \, x_{j} > v \} \, .
	\end{align*}
	The sets $N_1$ and $N_2$ then specify new rules $R_1$ and $R_2$, respectively. Both $x_j$ and $v$ are chosen so as to achieve a maximal  reduction of variance
	\begin{align}
	Var(N) - \left( \frac{|N_1|}{|N|}  Var(N_1) + \frac{|N_2|}{|N|} Var(N_2) \right)
	\enspace ,
	\label{Methods:eq:SDR}
	\end{align}
	where $Var(N)$ is the variance of the target attribute (the $y$-values) of the instances in $N$.
	
	Variance reduction has its roots in the earliest decision tree induction methods, in which  splits are chosen that decrease the impurity of leaf nodes. For categorical target attributes,  this is usually put in practice by reducing the information entropy. In the case of classification, the majority class is then used for prediction at a leaf node.
	In regression, where the target attribute is numerical, averaging is a more reasonable aggregation strategy; it was already adopted by the first regression tree learner CART \citep{Breiman:1984:CRT:CART}. 
	With the aim of minimizing the sum of squared errors, variance reduction becomes the right splitting criterion, since the sum of weighted variances (the second part of (\ref{Methods:eq:SDR})) can be written as the sum of squared errors:
	\begin{equation}
	\sum_{l=1}^2{ \frac{|N_l|}{|N|}  Var(N_l)}  = { \frac{1}{|N|}} \sum_{(\vec{x}_i,y_i) \in N_1}{ \big(y_i- \bar{y}(N_1) \big)^2}  
	 + { \frac{1}{|N|}} \sum_{(\vec{x}_i,y_i) \in N_2} { \big(y_i - \bar{y}(N_2) \big)^2} \enspace ,
	\label{Methods:eq:SDR1}
	\end{equation}
	where $\bar{y}(N_l) = \frac{1}{|N_l|} \sum_{(\vec{x}_i,y_i) \in N_l} y_i$ is the (constant) prediction produced by the rule $R_l$.
	
	M5 \citep{Quinlan:1992:LWCC:M5}, one of the most popular regression approaches, is a tree that is similar to regression trees with the exception of learning a linear function in the leaf nodes, instead of predicting a constant (the average in CART), while employing variance reduction as a splitting criterion. FIMTDD extends M5 for learning model trees from data streams; it also applies variance reduction as a splitting criterion.
	
	Despite the popularity of variance reduction, it has been criticized by \cite{Karalic:1992:ELRRTL} as ``not an appropriate measure for impurity of an example set since example sets with large variance and very low impurity can arise''. Similarly, a set of data points might be perfectly located on a hyperplane, non-orthogonal to the target axis, and still have a high variance.

	FLEXFIS and eTS+ do not apply a splitting criterion directly, but utilize an extension mechanism that decides when to add rules to the current rule set. More specifically, FLEXFIS applies an incremental clustering method, namely an incremental version of vector quantization \citep{Gray:1984:VQ}, such that a new example forms a new cluster if its distance to the nearest cluster is larger than the ``vigilance'' parameter. This parameter controls the tradeoff between major structural changes (creating a new cluster) and minor adaptations of the current structure.
	Likewise, eTS+ utilizes a density-based incremental clustering, eClusteting+ \citep{Angelov:2004:AAFRBAUOC}. In both approaches, the clusters found are eventually transformed into rules.

	Finally, we mention that most of the presented approaches consider only a single attribute for splitting, which leads to axis-parallel splits, not only in the standard case (FIMTDD and AMRules) but also in the case of fuzzy methods. FLEXFIS and eTS+ constitute an exception, since they find multivariate Gaussian clusters with non-diagonal covariance matrices.
	
	\subsection{Statistical tests versus engineered parameters}
	Learning on data streams, including the choice of the next split, must be done in an online manner. To answer the question whether or not an additional split is required, i.e., whether or not a significant improvement can be achieved through a split, statistical tests can be applied. A statistical test based on the Hoeffding bound has been extensively used by recent machine learning approaches for classification and regression, including Hoeffding trees, FIMTDD, AMRules, and TSK-Streams.
	
	Instead of applying statistical tests, FLEXFIS and eTS+ make use of more engineered solutions, such as creating a new rule whenever an example is distant from all existing rules, as in FLEXFIS, or when adding an example reduces the density of existing ones, as in eTS+.
	
%
%
%

	\section{The Learning Algorithm TSK-Streams}
	
	TSK-Streams is an incremental, adaptive algorithm for learning rule-based regression models in a streaming mode. More specifically, TSK-Streams produces a widely used type of fuzzy rule system called Takagi-Sugeno-Kang (TSK) \citep{Takagi:1985:FISAM}.
		
	\subsection{TSK Fuzzy Systems}
	
	A TSK rule $R_i$ has the following structure: 
	\begin{align}
	& \text{IF} \; \quad (x_1 \hspace{0.1cm} \text{IS}
	\hspace{0.1cm} A_{i,1}) \hspace{0.2cm} \text{AND}
	\quad  \ldots \quad \text{AND} \hspace{0.2cm} (x_d \hspace{0.1cm} \text{IS}
	\hspace{0.1cm} A_{i,d}) \nonumber   \\
	& \text{THEN} \quad l_i(\vec{x}) = w_{i,0}+w_{i,1}x_1+ \ldots +w_{i,d}x_d \enspace ,
	\label{Methods:eq:TSK_Rule}
	\end{align}
	with $(x_1, \ldots , x_d)^\top$ the representation of an instance $\vec{x} \in \mathbb{R}^d$ in terms of feature values, and $A_{i,j}$ the $j^{th}$ antecedent of $R_i$ in terms of a soft constraint. The consequent part of the rule is specified by the vector $\vec{\omega} = (w_{i,0}, \ldots , w_{i,d})^\top \in \mathbb{R}^{d+1}$, which defines an affine function of the input features. 
	In what follows, we denote a rule by $R_i=(M_i,\vec{\omega}_i)$, with 
	$M_i$ the fuzzy sets defining the rule antecedents, and $\vec{\omega}_i$ the coefficients specifying the linear function.
	
	A fuzzy set with membership function $\mu_{j}^{(i)}: \, \mathbb{R} \longrightarrow [0,1]$ is used to model the soft constraint $A_{i,j}$. Thus, the predicate $(x_j \hspace{0.1cm} \text{IS} \hspace{0.1cm} A_{i,j})$ has truth degree $\mu_{j}^{(i)}(x_j)$, which corresponds to the membership degree of $x_j$ in $\mu_{j}^{(i)}$. The overall degree to which an instance $\vec{x}$ satisfies the rule premise $R_i$ is 
	\begin{equation}\label{eq:ta}
	\mu_i(\vec{x}) = \top\left( \mu_{1}^{(i)}(x_1) , \ldots  , \mu_{d}^{(i)}(x_d) \right)  \, ,
	\end{equation}
	where the triangular norm\footnote{A triangular norm is a binary operator $\top:\, [0,1]^2 \longrightarrow [0,1]$, which is commutative, associative, non-decreasing in both arguments, and with neutral element $1$ and absorbing element $0$.} $\top$ models the logical conjunction \citep{Klement:2000:TN}. 
	We will adopt the G\"odel t-norm, which is given by $\top(u,v)=\text{min}(u,v)$. Notice that $A_{i,j}$ might be a void constraint, which corresponds to setting $\mu_{j}^{(i)} = \mu_{\text{void}} \equiv 1$;
	in that case, the feature $x_j$ is effectively removed from the premise of the rule (\ref{Methods:eq:TSK_Rule}).

	Now, given an instance $\vec{x}$ as an input to a TSK system with $C$ rules $RS = \{ R_1, \ldots , R_C \}$, each of these rules will be ``activated'' with the degree (\ref{eq:ta}). Therefore, the system's output is specified by the weighted average of the outputs suggested by the individual rules:
	\begin{equation}
	\label{Methods:eq:TSK_Output}
	\hat{y} = \sum_{i=1}^{C} \Psi_i(\vec{x}) \cdot l_i(\vec{x}) \enspace ,
	\end{equation}
	with
	\begin{equation}
	\label{Methods:eq:TSK_Weighted_Output}
	\Psi_i(\vec{x}) =
	\frac{\mu_i(\vec{x})}{\sum_{j=1}^{C} \mu_j(\vec{x})}  \enspace .
	\end{equation}
	
	Fuzzy sets can be characterized by any function of the form $\mu: X \longrightarrow [0,1]$, which leads to membership functions with different shapes and properties \citep{Pedrycz:1998:AITFSAD}. In our approach, we employ the family of the ``S-shaped'' parametrized functions: a fuzzy set $\mu$ has a support and core $[a,d]$ and $[b,c]$, respectively, such that $[a,d] \supset [b,c]$, the degree of membership is 1 inside $[b,c]$ and $0$ outside $[a,b]$. The left boundary of the fuzzy set $\mu$ is modeled in terms of an ``S-shaped'' transition between zero and full membership:
	
	\begin{equation}\label{eq:S-Shaped}
	\mu_{\text{S}}(x)= \left\{ \begin{array}{cl}
	0 & \text{ if } x < a \\
	2 \left( \frac{x-a}{b-a} \right)^2 & \text{ if } a \leq x < \frac{a+b}{2} \\
	1 - 2 \left( \frac{x-b}{b-a} \right)^2 & \text{ if } \frac{a+b}{2} \leq x  < b \\
	1 & \text{ if } b \leq x \leq c \\
	1 - 2 \left( \frac{c-x}{d-c} \right)^2 & \text{ if } c < x \leq \frac{c+d}{2} \\
	2 \left( \frac{d-x}{d-c} \right)^2 & \text{ if } \frac{c+d}{2} < x \leq d \\
	0 & \text{ if } x > d  
	\end{array} \right.
	\enspace .
	\end{equation}
	An S-shaped membership function can also be left- or right-unbounded:
	\begin{equation}\label{eq:Left-unbounded}
	\mu_{\text{left-ub}}(x)= \left\{ \begin{array}{cl}
	1 & \text{ if } x < a \\
	1 - 2 \left( \frac{a-x}{b-a} \right)^2 & \text{ if } a < x \leq \frac{a+b}{2} \\
	2 \left( \frac{b-x}{b-a} \right)^2 & \text{ if } \frac{a+b}{2} < x \leq b \\
	0 & \text{ if } x > b  
	\end{array} \right.
	\enspace ,
	\end{equation}
	\begin{equation}\label{eq:Right-unbounded}
	\mu_{\text{right-ub}}(x)= \left\{ \begin{array}{cl}
	0 & \text{ if } x < a \\
	2 \left( \frac{x-a}{b-a} \right)^2 & \text{ if } a \leq x < \frac{a+b}{2} \\
	1 - 2 \left( \frac{x-b}{b-a} \right)^2 & \text{ if } \frac{a+b}{2} \leq x  < b \\
	1 & \text{ if } b \leq x \\
	\end{array} \right.
	\enspace .
	\end{equation}
	
	\subsection{Online Rule Induction}

	The TSK-Streams algorithm begins with a single default rule and then learns rules in an incremental manner. The default rule has an empty premise for each feature (that is, the membership function $\mu_{\text{void}}$) and covers the complete input space. 
	Then, the algorithm continuously checks whether, for any of the rules $R_i$, one of its extensions could possibly improve the performance of the current fuzzy system. 
	
	An expansion of a rule $R_i$ with a predicate $(x_j \text{ IS } A_{i,j})$ on the $j^{th}$ attribute means that the rule is split into two new rules $R_i^{\prime}$ and $R_i^{\prime\prime}$ with predicates 
	$(x_j \text{ IS } A^{\prime}_{i,j})$ and $(x_j \text{ IS } A^{\prime\prime}_{i,j})$,
	respectively, 
	where $ A_{i,j} = A^{\prime}_{i,j} \cup A^{\prime\prime}_{i,j}$.
	We denote the membership functions modeling the fuzzy sets $A^{\prime}_{i,j}$ and $A^{\prime\prime}_{i,j}$ by $\mu_{j}^{\prime(i)}$ and $\mu_{j}^{\prime\prime(i)}$, respectively.
	These membership functions are chosen after a fuzzy partitioning of the domain of feature $x_j$. To this end, we apply a supervised discretization technique that divides a fuzzy set into two new fuzzy sets so as to improve the overall performance. Here, we focus on two criteria (cf.\ the discussion in Section \ref{Learning Regression Models on Data Streams}), to be detailed in the following.
	
	\subsubsection{Variance reduction} 
	Similar to the AMRule principle of reducing the variance, based on the fuzzy set $A_{i,j}$, two fuzzy sets $A^{\prime}_{i,j}$ and $A^{\prime\prime}_{i,j}$ are created such that a maximum reduction in the target attribute's variance is achieved.
	For example, let $A_{i,j}$ be a fuzzy set (for the $j^{th}$ attribute in the rule $R_i$) characterized by the S-shaped membership function $\mu_{j}^{(i)}$, which is parametrized by the quadruple $(a,b,c,d)$. Let $N_{R_i}$ be the set of examples $(\vec{x},y)$  covered by the rule $R_i$, i.e., the examples for which $\mu_i(\vec{x})>0$. We then seek to find the value $q \in [a,d]$ such that the reduction in variance 
	$$
	Var(S) - \big(
	w'  Var(S^{\prime}) 
	+ w'' Var(S^{\prime\prime}) \big)
	$$
	is maximal,	where
	\begin{align*}
	S & =\{y \cdot \Psi_i(\vec{x}) \given  ( \vec{x},y ) \in N_{R_i}\} , \\ 
	S^{\prime}& =\{y \cdot \Psi_i(\vec{x}) \given (\vec{x},y) \in N^{\prime}_{R_i} \} , \\
	S^{\prime\prime} & =\{y \cdot \Psi_i(\vec{x}) \given (\vec{x},y) \in N^{\prime\prime}_{R_i} \}, \\
	w' & = \frac{\sum_{(\vec{x},y) \in N^{\prime}_{R_i}} \Psi_i(\vec{x}) }{\sum_{(\vec{x},y) \in N_{R_i}} \Psi_i(\vec{x})} , \\
	w'' & = \frac{\sum_{(\vec{x},y) \in N^{\prime\prime}_{R_i}} \Psi_i(\vec{x}) }{\sum_{(\vec{x},y) \in N_{R_i}} \Psi_i(\vec{x})} ,
	\end{align*}
	with 
	\begin{align*}
	N^{\prime}_{R_i} & = \{(\vec{x},y) \given (\vec{x},y) \in N_{R_i} , \, x_j \in [a,q] \} ,\\
	N^{\prime\prime}_{R_i} & = \{(\vec{x},y) \given (\vec{x},y) \in N_{R_i} , \,  x_j \in [q,d] \} ,
	\end{align*}
	and $Var(S)$ the variance of the set $S$.
	 
	Similar to AMRules and FIMTDD, we achieve the variance reduction by storing candidate values in an extended binary search tree (E-BST). This data structure allows for computing the variance reduction for each candidate value in time that is linear in the size of the tree; moreover, it can be updated in logarithmic time \citep{ikonomovska2012algorithms}. 
	
	\subsubsection{Error reduction} 
	Extending the current model with new rules so as to improve the system's overall performance requires, for each existing rule, the creation and evaluation of all possible extensions---evaluating an extension here means determining the empirical performance of the (modified) system as a whole. As before, by a possible rule extension we mean replacing a fuzzy set $A_{i,j}$ in a rule antecedent by new fuzzy sets $A_{i,j}'$ and $A_{i,j}''$, which are produced by bisecting the support of $A_{i,j}$ at some suitable splitting point. Even if these splitting points were organized in a binary search tree structure, the number of updates required after observing a new example would no longer be logarithmic but linear. Indeed, every possible extension means fitting a step-wise linear function, at each splitting value, on the entire training data (or updating the linear function on the new data instance).
	
	To counter the aforementioned problem, we suggest a heuristic that simultaneously chooses a promising splitting value and fits a stepwise linear function for each candidate extension rule. The splitting value is chosen by adaptively shifting (increasing or decreasing) it based on the performance of new candidate rules.
	More formally, let $A_{i,j}$ be a fuzzy set characterized by the S-shaped membership function $\mu_{j}^{(i)}$, parametrized by $(a,b,c,d)$, and let $N_{R_i}$  be the set of instances $(\vec{x},y)$ covered by the rule $R_i$. Let $q \in [a,d]$ be the initial splitting value from which $A^{\prime}_{i,j}$ and $A^{\prime\prime}_{i,j}$ are constructed via suitable parametrizations $(a,b,q+\rho_1,q+\rho_2)$ and $(q-\rho_2,q-\rho_1,c,d)$ of their membership functions $\mu_{j}^{\prime(i)}$ and $\mu_{j}^{\prime\prime(i)}$, respectively.
	We initialize $q$ by the current mean of the observed values $x_j$. The values $\rho_1$ and $\rho_2$ control the steepness of the S-shaped function and are chosen in proportion to the observed variance. From the membership functions $\mu_{j}^{\prime(i)}$ and $\mu_{j}^{\prime\prime(i)}$ and the parent rule $R_i$, the new candidate rules $R^{\prime}_i$ and $R^{\prime\prime}_i$ are created (see lines 1--9 of Algorithm \ref{alg.GenUpdateERCandidates}).
	
	Upon observing a new example $(\vec{x},y)$, both the membership degrees $\mu_{j}^{\prime(i)}(\vec{x})$,	$\mu_{j}^{\prime\prime(i)}(\vec{x})$ and the errors committed by each candidate rule, $err_1 = (\vec{\omega}^{\prime(i)} \cdot \vec{x} - y)^2$, $err_2 = (\vec{\omega}^{\prime\prime(i)} \cdot \vec{x} - y)^2$, are computed.
	If the ``winner rule'', i.e., the candidate rule by which the example is covered the most, commits an error that is larger than the error committed by the other candidate rule (covering the example to a lesser degree), we consider this as an inconsistency. The latter can be mitigated by shifting the splitting value $q$ right or left, in proportion to the error committed by each candidate extension (see lines 11--21 of Algorithm \ref{alg.GenUpdateERCandidates}).

	\begin{algorithm}
		\caption{GenUpdateERCandidates -- ErrorReduction}
		\label{alg.GenUpdateERCandidates}
		\KwIn{
			$R_i, S_i, (\vec{x}_t,y_t)$:\\
			$R_i=(M_i,\vec{\omega}_i)$: the rule whose extensions should be created/updated \\
			$M_i$: the set of fuzzy sets conjugated in the premise.\\
			$\vec{\omega}_i$: the vector of coefficients of the linear function.\\
			$S_i=\{(R^{\prime}_i,R^{\prime\prime}_i)\}$: Set of candidate extensions of rule $R_i$. \\
			$(\vec{x}_t,y_t)$: a new training example.
		}
		\If{$S_i$ \text{is Empty}}
		{
			\For{ $j \in \{1,\dots,d\}$}
			{
				Update $Mean(x_{tj}), Var(x_{tj})$ \\
				$M^{\prime}_i = \{\mu_{j}^{\prime(i)} = (a,b,q+\rho_1,q+\rho_2)\} \cup M_i \setminus \{ \mu_{j}^{(i)} \} $\\
				$M^{\prime\prime}_i = \{\mu_{j}^{\prime\prime(i)} = (q-\rho_2,q-\rho_1,c,d)\} $\\ 
				$\quad\quad\quad\cup M_i \setminus \{ \mu_{j}^{(i)} \} $\\
				$\vec{\omega}^{\prime}_i = \vec{\omega}_i, \quad\quad \vec{\omega}^{\prime\prime}_i = \vec{\omega}_i$\\
				$R^{\prime}_i=(M^{\prime}_i, \vec{\omega}^{\prime}_i), \quad\quad $			
				$R^{\prime\prime}_i=(M^{\prime\prime}_i,\vec{\omega}^{\prime\prime}_i)$\\
				$S_i = S_i \cup  \{s_j= (R^{\prime}_i,R^{\prime\prime}_i)\}$ \\
			}
		}
		\Else{
			\For{ $j \in \{1,\dots,d\}$}
			{	
				Find $s_j = (R^{\prime}_i,R^{\prime\prime}_i) \in S_i$ s.t. $R^{\prime}_i=(M^{\prime}_i,\vec{\omega}^{\prime}_i),   R^{\prime\prime}_i=(M^{\prime\prime}_i,\vec{\omega}^{\prime\prime}_i)  \wedge  M_i \setminus \{ M^{\prime}_i \} = M_i \setminus \{ M^{\prime\prime}_i \} =\{\mu_{j}^{(i)}\} $\\
				$m_1 = \mu_{j}^{\prime(i)}(x_{tj}), \quad error_1 = (\vec{\omega}^{\prime}_i \cdot \vec{x}_t - y_t)^2$\\
				$m_2 = \mu_{j}^{\prime\prime(i)}(x_{tj}), \quad error_2 = (\vec{\omega}^{\prime\prime}_i \cdot \vec{x}_t - y_t)^2$\\
				\If{$m_1>m_2 \wedge error_1 > error_2$}
				{
					\tcc{shift $q$ to the left}
					$q = q - \eta \Psi_i(\vec{x}_t) (error_1 - error_2)$ 		
				}
				\ElseIf{$m_1<m_2 \wedge error_1 < error_2$}
				{
					\tcc{shift $q$ to the right}
					$q = q - \eta \Psi_i(\vec{x}_t) (error_1 - error_2)$
				}
				Update $\mu_{j}^{\prime(i)} = (a,b,q+\rho_1,q+\rho_2)\}$ \\
				Update $\mu_{j}^{\prime\prime(i)} = (q-\rho_2,q-\rho_1,c,d)\}$ \\
			}
			\tcc{Update $\vec{\omega}^{\prime}_i, \vec{\omega}^{\prime\prime}_i$, Algorithm \ref{alg.UpdateConsequent}} 
			UpdateConsequent($R_i,S_i$) \\
		}
	\end{algorithm}
	
	In the explanations above, we outlined two ways of splitting an S-shaped function into two such  functions of similar shape. In the beginning, however, the default rule contains only unbounded fuzzy sets characterized by $\mu_{\text{void}}$. A split of an unbounded fuzzy set produces two sets with membership functions $\mu_{\text{left-ub}}(x)$ and $\mu_{\text{right-ub}}(x)$, respectively, which cover the resulting half spaces (with some degree of overlap).
	Similarly, a split of a right- or left-unbounded membership function leads to a right- or left-unbounded and an S-shaped function.
	
	Recall that AMRules adopts only a single rule from the two candidates emerging from a rule expansion (cf.\ Section \ref{Trees versus rules}). More specifically, AMRules keeps the rule with minimum weighted variance and discards the other candidate as well as the parent rule from the original rule set. Since the resulting rule set does not form a partition of the instance space, this strategy requires a default rule covering the space that is not covered by any other rule. Motivated by this strategy, we also study the effect of adopting only a single instead of both rule extensions. Thus, we distinguish the following two strategies.
	\begin{enumerate}
		\item Single Extension: Only the best extension is added to the rule set, while the other one is discarded. The parent rule is also discarded unless it is the default rule. The choice of the best rule depends on the criterion used for splitting: either the weighted variance reduction or the weighted SSE.	
		\item All Extensions: Both extensions are added to the rule set, and the parent rule is removed. This approach makes the whole system of rules equivalent to a tree structure.	
	\end{enumerate}
	The two adaptation strategies will be revisited in the context of change detection in Section~\ref{Change Detection}. A more detailed exposition of the adaptation strategies is given in Algorithms \ref{alg.ExpandSystemVarianceReduction} and \ref{alg.ExpandSystemErrorReduction}.

	\subsection{Rule Consequents}
	
	FLEXFIS makes use of recursive weighted least squares estimation (RWLS) \citep{Ljung:1999:SITU} to fit linear functions as rule consequents. This approach is computationally expensive, as it requires multiple matrix inversions. 	
	In our approach, and similar to AMRules, we learn consequents more efficiently using gradient methods.

	When a new training instance $(\vec{x}_t,y_t)$ arrives, TSK-Streams produces a prediction $\hat{y}_t$, the squared  error of which can be obtained as follows:
	\begin{align}
	E_t &= (y_t - \hat{y}_t)^2 \\
	&= \left(y_t- \left(\sum_{R_i \in RS} \frac{\Psi_{i}(\vec{x}_t)}{\sum_{R_k \in RS} \Psi_k(\vec{x}_t) } \sum_{j=0}^{d} \omega_{i,j} x_{t,j} \right)\right)^2 ,
	\end{align}
	where $RS$ is the current set of rules.
	According to the technique of stochastic gradient descent, the coefficients $\omega_{i,j}$ are then moved into the negative direction of the gradient, with the length of the shift being controlled by the learning rate $\eta$:
	\begin{equation}
	\vec{\omega} \leftarrow \vec{\omega} - \eta \nabla E_t \enspace  .
	\end{equation} 
	Thus, the following (component-wise) update rule is obtained:
	\begin{equation*}
	\omega_{i,j} \leftarrow \omega_{i,j} - 2\, \eta  (y_t - \hat{y}_t) \left(\sum_{R_i \in RS} \frac{\Psi_i(\vec{x}_t)}{\sum_{R_k \in RS} \Psi_k(\vec{x}_t) } \,  x_{t,j} \right)
	\end{equation*}
	The process of updating the rule consequents is summarized in Algorithm \ref{alg.UpdateConsequent}, which also updates the consequents of the rule's extension (when the error reduction strategy is used).
	
	\begin{algorithm}
		\caption{UpdateConsequent}
		\label{alg.UpdateConsequent}
		\KwIn
		{
			$RS=\{(R, S)\}, R_i, S_i, (\vec{x}_t, y_t)$\\
			$RS=\{(R, S)\}$: the set of all rules and their extensions. \\
			$R_i=(M_i,\vec{\omega}_i)$: the rule whose consequent and the consequents of its extensions should be updated. \\
			$M_i$: the set of fuzzy sets conjugated in premise.\\
			$\vec{\omega}_i$: the vector of coefficients of the linear function.\\
			$S_i=\{(R^{\prime}_i,R^{\prime\prime}_i)\}$: Set of candidate extensions of rule $R_i$. \\
			$(\vec{x}_t,y_t)$: a new training example
			
				}
		$m_1 = \sum_{R_j \in RS} \mu_j(\vec{x}_t)$\\
		$m_2 = \sum_{R_j \in RS} \mu_j(\vec{x}_t) l_j(\vec{x}_t)$\\
		$\mu_i(\vec{x}_t) = \top(\bigotimes_{\mu_{j}^{(i)} \in M_i} \mu_{j}^{(i)}(x_{tj}))$\\
		\If{$\mu_i(\vec{x}_t) > 0$}
		{
			\For{$(R^{\prime}_i,R^{\prime\prime}_i) \in S_i$}
			{
				$\mu^{\prime}_i(\vec{x}_t) = \top(\bigotimes_{\mu_{j}^{\prime(i)} \in M^{\prime}_i} \mu_{j}^{\prime(i)}(x_{tj}))$\\
				$\mu^{\prime\prime}_i(\vec{x}_t) = \top(\bigotimes_{\mu_{j}^{\prime\prime(i)} \in M^{\prime\prime}_i} \mu_{j}^{\prime\prime(i)}(x_{tj}))$\\

				${m_1}\prime = m_1 - \mu_i(\vec{x}_t) + \mu^{\prime}_i(\vec{x}_t) + \mu^{\prime\prime}_i(\vec{x}_t) $ \\
				${m_2}\prime = m_2  - \mu_i(\vec{x}_t) l_{i}(\vec{x}_t) + \mu^{\prime}_i(\vec{x}_t) l^{\prime}_i(\vec{x}_t) + \mu^{\prime\prime}_i(\vec{x}_t) l^{\prime\prime}_i(\vec{x}_t)$\\
				
				$\vec{\omega}^{\prime}_i = \vec{\omega}^{\prime}_i + \eta  (y_t-\frac{m_2\prime}{m_1\prime})
				\left(\frac{\mu^{\prime}_i(\vec{x}_t)}{m_1\prime} \vec{x}_t\right)$ \\
				$\vec{\omega}^{\prime\prime}_i = \vec{\omega}^{\prime\prime}_i + \eta (y_t-\frac{m_2\prime}{m_1\prime})
				\left(\frac{\mu^{\prime\prime}_i(\vec{x}_t)}{m_1\prime} \vec{x}_t\right)$ \\
			}
			$\vec{\omega}_i = \vec{\omega}_i + \eta (y_t-\frac{m_2}{m_1})
			\left(\frac{\mu_i(\vec{x}_t)}{m_1} \vec{x}_t\right) $ \\  
		}
	\end{algorithm}
	
	\subsection{Model Structure}
	
	TSK-Streams adapts the TSK rule system (that is, the fuzzy sets in the rule antecedents and the linear function in the consequents) in a continuous manner. While the  adaptations discussed so far essentially concern the parameters of the system, the replacement of a rule by one of its expansions corresponds to a (more substantial) structural change. 
	
	For obvious reasons, such changes should be handled with caution, especially when they lead to an increased complexity of the model. Learning methods therefore tend to maintain the current model unless being sufficiently convinced that an expansion will yield an improvement. To decide whether or not a possible expansion should be adopted, the estimated performance difference is typically taken as a criterion: this difference should be \emph{significant} in a statistical sense. 
	
	
	In our algorithm, we make use of Hoeffding's inequality to support these decisions. 
	The latter bounds the difference between the empirical mean $\bar{X}$ of the $n$ i.i.d.\ random variables 
	$X_1, \ldots , X_n$ (having support $[a,b] \subset \mathbb{R}$) and the expectation $E(\overline{X})$ in terms of
	\begin{equation}\label{eq:hi}
	P \Big( \vert \bar{X} -\mathrm{E} (X) \vert > \epsilon \Big) \leq \exp \left(-{\frac {2n\epsilon^{2}}{(b-a)^{2}}}\right) \enspace .
	\end{equation}
	More specifically, when using the error reduction criterion, we replace a rule $R_i$ by two rules $R^{\prime}_i$ and $R^{\prime\prime}_i$, considering the reduction in the sum of squared errors (SSE). That is, the SSE of the current rule set $RS$ is compared with the SSE of all alternative systems ($RS \setminus R_i) \cup \{ R^{\prime}_i, R^{\prime\prime}_i \}$. With $SSE_{best}$ and $SSE_{2ndbest}$ denoting the expansion with the lowest and the second lowest error, respectively, the best expansion is adopted if 
	\begin{equation}\label{eq:ra}
	\frac{SSE_{best}}{SSE_{2ndbest}} < 1  - \epsilon \enspace ,
	\end{equation}
	or when $\epsilon$ falls below a tie-breaking constant $\tau$. The constant $\epsilon$ is obtained from (\ref{eq:hi}) by setting the probability to a desired degree of confidence $1- \delta$, i.e., setting the right-hand side to $1-\delta$ and solving for $\epsilon$; noting that the ratio (\ref{eq:ra}) is bounded in $]0,1]$, $b-a$ is set to 1. Algorithm \ref{alg.ExpandSystemVarianceReduction} depicts the system expansion procedure when the error reduction criterion is applied. The same technique can be used for the single extension variant, except that the rule $R_i$ is replaced with the extension that achieves the lowest weighted SSE (provided $R_i$ is not the default rule, otherwise $R_i$ is also kept).

	As an alternative to the global error reduction criterion, the variance reduction approach checks for the decrease in variance for each rule locally. The Hoeffding inequality is then applied to the ratio of the variance reductions of the best two candidate extensions of the same rule $R_i$. The procedure that performs the expansion is depicted in Algorithm \ref{alg.ExpandSystemErrorReduction}. This strategy can be seen as a model adaptation through local improvements.
	
		\begin{algorithm}
		\caption{ExpandSystemVR -- VarianceReduction}
		\label{alg.ExpandSystemVarianceReduction}
		\KwIn{$RS=\{(R, S)\}, \delta, \tau, n$  \\
			$RS=\{(R, S)\}= \bigcup\limits_{R_i} \{(R_i, S_i)\} $: rules and extensions\\ 
			$\quad\quad = \bigcup\limits_{R_i} \{(R_i, \bigcup\limits_{j=1}^{d} \{s_j = (R^{\prime}_i,R^{\prime\prime}_i,VarRed_{ij})\}))\}$ \\
			$VarRed_{ij}$: variance reduction caused by the extension $j$ of rule $R_i$ \\
			$\delta$: confidence level \\
			$\tau$: tie-breaking constant\\
			$n$: number of examples seen by the current system\\
			$(\vec{x}_t,y_t)$: a new training example.
		}
		
		\For{$(R_i,S_i) \in RS$}
		{
			\textbf{let} $s_{best}=(R^{\prime}_i,R^{\prime\prime}_i,VarRed_{best}) \in S_i$ has the largest VarRed\\
			\textbf{let} $s_{2ndbest} =(R^{\prime}_i,R^{\prime\prime}_i,Red_{2ndbest}) \in S$ has the 2nd largest VarRed\\
			$\epsilon = \sqrt{\frac{\ln\left(\frac{1}{\delta}\right)\left(R\right)^{2} }{2n}}$ + complexity\\
			$\overline{X} = \frac{VarRed_{2ndbest}}{VarRed_{best}}$ \\
			\If{$((\overline{X} + \epsilon) < 1$ OR $\epsilon < \tau)$}
			{
				\If{Single Extension}
				{
					let $R_{best} \in \{R^{\prime}_i,R^{\prime\prime}_i\}$ has the largest weighted VarRed\\
					$RS = RS \cup \{(R_{best},GenerateExtendedRules(R_{best})\}$ \\
					\If{$R_i$ \text{is not} $R_{default}$}
					{
						$RS = RS \setminus \{(R_i,S_i)\} $ \\	
					}		
				}
				\Else
				{
					$RS = RS  \cup  \{(R^{\prime}_i,GenerateExtendedRules(R^{\prime}_i), $ \\ $\quad\quad(R^{\prime\prime}_i,GenerateExtendedRules(R^{\prime\prime}_i)\}$ \\
					$RS = RS \setminus \{(R_i,S_i)\} $ \\	
				}
			}	
			
		}
	\end{algorithm}
	
	\begin{algorithm}
		\caption{ExpandSystemER -- ErrorReduction}
		\label{alg.ExpandSystemErrorReduction}
		\KwIn{$RS=\{(R, S)\}, \delta, \tau, n$  \\
			$RS=\{(R, S)\}= \bigcup\limits_{R_i} \{(R_i, S_i)\} $: rules and extensions\\ 
			$\quad\quad = \bigcup\limits_{R_i} \{(R_i, \bigcup\limits_{j=1}^{d} \{s_j = (R^{\prime}_i,R^{\prime\prime}_i,SSE_{ij})\}))\}$ \\
			$SSE_{ij}$: the sum of squared errors committed by the extension $j$ of rule $R_i$ \\
			$\delta$: confidence level \\
			$\tau$: tie-breaking constant\\
			$n$: number of examples seen by the current system\\
			$(\vec{x}_t,y_t)$: a new training example.
		}
		
		\textbf{let} $SSE_{current}$ be the SSE of the current system\\
		\textbf{let} $s_{pq}$ be the extension with smallest SSE \\
		\textbf{let} $s_{pq}$ be the extension with second smallest SSE \\
		Update $SSE_{current}$, $s_{pq}$ and $s_{pq}$ on $(\vec{x}_t,y_t)$
		$\epsilon = \sqrt{\frac{\ln\left(\frac{1}{\delta}\right)\left(R\right)^{2} }{2n}}$ + complexity\\

		$\overline{X} = \frac{1}{n} (SSE_{pq} / SSE_{uv})$ \\
		$\overline{Y} = \frac{1}{n} (SSE_{pq} / SSE_{current})$ \\
		
		\If{$( (\overline{Y} + \epsilon) < 1)$ AND $((\overline{X} + \epsilon) < 1$ OR $\epsilon < \tau)$}
		{
			\If{Single Extension}
			{
				let $R_{best} \in \{R^{\prime}_p,R^{\prime\prime}_p\}$ has the smallest weighted SSE\\
				$RS = RS \cup \{(R_{best},GenUpdateERCandidates(R_{best},\emptyset,(\vec{x}_t,y_t))\}$ \\
				\If{$R_p$ \text{is not} $R_{default}$}
				{
					$RS = RS \setminus \{(R_p,S_p)\} $ \\	
				}		
			}
			\Else
			{
				$RS = RS  \cup  \{(R^{\prime}_p,GenUpdateERCandidates(R^{\prime}_p,\emptyset,(\vec{x}_t,y_t)), $ \\ $\quad\quad(R^{\prime\prime}_p,GenUpdateERCandidates(R^{\prime\prime}_p,\emptyset,(\vec{x}_t,y_t))\}$ \\
				$RS = RS \setminus \{(R_p,S_p)\} $ \\	
			}
		}
	\end{algorithm}

	Finally, we propose a penalization mechanism to avoid a danger of overfitting due to an excessive increase of the number of rules. This mechanism consists of adding a complexity term $C$ to $\epsilon$. For both extensions (variance reduction and error reduction), $C$ is set to $d^{-2}\sqrt{|RS|}$, with $d$ the number of features and $RS$ the current rule set.

	\subsection{Change Detection}
	\label{Change Detection}

	A concept drift may cause a drop in the performance of a rule. To detect such cases, we make use of the adaptive windowing (ADWIN) \citep{Bifet:2007:LFTCDWAW} drift detector. Compared to the Page-Hinkely test (PH) \citep{PAGE:1954:CIS}, which is used by AMRules, ADWIN has the advantage of being non-parametric, which means that it makes no assumptions about the observed random variable. Besides, only a single parameter needs to be chosen, namely the tolerance towards false alarms ($\delta_{adwin}$). In our approach, ADWIN is locally applied in each rule. More specifically, given that an example is covered by a rule, it is applied on the absolute error committed by that rule on this example. 
	
	For the single extension strategy, the rule that suffers from a drop of performance can be simply discarded. But in the all extensions strategy and upon detecting a drift in the rule $R_p=(M_p,\vec{\omega}_p)$, we find its sibling rule $R_q=(M_q,\vec{\omega}_q)$, from which it differs by only one single literal
	(i.e., there is a fuzzy set $\mu_{j}^{(p)} \in M_p$ on the $j^{th}$ attribute that satisfies the following criterion: 
	for all $i \in \{1,  \ldots , d \} \setminus \{j\}: \mu_{i}^{(p)} \in M_q \wedge \mu_{i}^{(q)} \in M_p$).
	To remove the rule $R_p$, it is retracted from the rule set, and its sibling rule $R_q$ is updated by replacing $\mu_{j}^{(q)}$ with $\mu_{j}^{(p)} \cup \mu_{j}^{(q)}$.
	In case the sibling rule $R_q$ has already been extended before the drift is detected, the same procedure is applied recursively to the children of this rule.

	\section{Empirical Evaluation}
	
	To compare our method TSK-Streams with existing algorithms, we conducted a series of experiments, in which we investigated the algorithms' predictive accuracy,  their runtime, and the size of the models they produce.

	\subsection{Methods, Data, and Experimental Setup}
	
	TSK-Streams is implemented in MOA\footnote{\url{http://moa.cms.waikato.ac.nz}} (Massive Online Analysis) \citep{Bifet:2010:MOAMOA}, which is an open source software framework for mining and analyzing large data sets in a streaming mode. In our experiments, TSK-Streams is compared with AMRules, FIMTDD, ARF-Reg, and FLEXFIS. Both AMRules and FIMTDD are implemented in MOA's distribution, and we use them in their default settings with $\delta=0.01$ and $\tau=0.05$ (the significance level of the Hoeffding inequality and the tie-breaking constant). 
	We implement ARF-Reg as described in the original paper \citep{Gomes:2018:ARFDSR} by setting $\lambda=6$, the ensemble size $L=10$, and the number of features $m =\sqrt{d} +1$, with $d$ being the total number of features. As for the parametrization of TSK-Streams,  maximal comparability with AMRules and FIMTDD is assured by using the same values $\delta$, $\tau$.	FLEXFIS is implemented in Matlab. Its parameters were tuned with the help of a function specifically offered for that purpose. The only exception is the ``forgetting parameter'', for the value $0.999$ was (manually) found to provide the best performance. 
	
	The test-then-train protocol was used for all experiments. According to this protocol, each instance is used for both testing and training: The model is evaluated on the instance first, and a learning step is carried out afterward. Experiments are performed on benchmark data sets\footnote{The first 14 data sets are the same as those used in \citep{Gama:2013:AMRFDS}.} collected from the UCI repository\footnote{\url{http://archive.ics.uci.edu/ml/}} \citep{Lichman:2013:UCI} and other repositories\footnote{\url{https://github.com/renatopp/arff-datasets/tree/master/regression }, \url{http://tunedit.org/repo/UCI/numeric}}; a summary of the type, the number of attributes and instances of each data set is given in Table \ref{tab:datasets}.
	
	The data sets starting with prefix \textit{BNG-} are obtained from the online machine learning platform OpenML \citep{BischlOpenML2017}; these large data streams are drawn from Bayesian networks as generative models, after constructing each network from a relatively small data set (we refer to \cite{van2014bayesian} for more details).
	
	\begin{table}[h]
		\caption{Data Sets}
		\label{tab:datasets}
		\centering
		\begin{tabular}{llcrr}
			\hline
			\# & Name  & Synthetic & Instances & Attributes \\
			\hline
			1&2dplanes&yes&40768&11\\
			2&ailerons&no&13750&41\\
			3&bank8FM&yes&8192&9\\
			4&calHousing&no&20640&8\\
			5&elevators&no&8752&19\\
			6&fried&yes&40769&11\\
			7&house16H&no&22784&16\\
			8&house8L&no&22784&8\\
			9&kin8nm&-&8192&9\\
			10&mvnumeric&yes&40768&10\\
			11&pol&no&15000&49\\
			12&puma32H&yes&8192&32\\
			13&puma8NH&yes&8192&9\\
			14&ratingssweetrs&-&17903&2\\
			15&BNG-stock&semi&59049&10\\
			16&BNG-cholesterol&semi&100000&14\\
			17&BNG-echoMonths&semi&17496&10\\
			18&BNG-wine-quality&semi&100000&14\\
			\hline
		\end{tabular}
	\end{table}
	
	\subsection{Results}

	In the first part of the evaluation, we compare the four variants of our own proposal: variance reduction versus error reduction, and the extension using a single candidate versus the extension for both candidates.

	Table \ref{tbl.Comparing.performanceTSK-Streamsvariants} shows the average RMSE and the corresponding standard error on ten rounds for each data set. In this table, the last row shows the number of wins/losses of the first three against the fourth variant (with variance reduction and consideration of both candidates); these tests apply the Wilcoxon signed-rank test over the paired performances of the 10 iterations with confidence level $\alpha = 0.05$. From the results, the fourth variant appears to be  superior to the other variants. Therefore, we adopt this variant (simply referred to as TSK-Streams in the following) and consider it for further comparison with state-of-the-art methods.
	
	Table \ref{tbl.Comparing.performanceVsVarianceReductionBothCand} presents the performance comparison between TSK-Streams and the other approaches, AMRules, FIMTDD, ARF-Reg, and FLEXFIS. Overall, TSK-Streams compares quite favourably and performs best in terms of the average rank statistic. Moreover, at least on 10 of the 18 data sets, its performance is statistically better (also according to the Wilcoxon signed-rank test at significance level $\alpha = 0.05$) than that of any other approach. With FIMTDD being the least performing method, its incremental random forest variant, ARF-Reg, presents a slightly better perfomance.	
	
	Other criteria important for the applicability of an approach in the setting of data streams include model complexity and efficiency. Obviously, these properties are not independent of each other, because more time is needed to maintain and adapt larger models. We measure the two criteria, respectively, in terms of the number of rules/leaf nodes in the model eventually produced by a learning algorithm and the average time (in milliseconds) the algorithms needs to process a single instance. We consider the latter more informative than the total runtime on an entire data set (stream), because the processing time per instance is more relevant for the possible application of an algorithm under real-time conditions. Table \ref{tbl.Comparing.SizeVsVarianceReductionBothCand} shows that TSK-Streams tends to produces smaller models than FIMTDD and ARF-Reg, which are still slightly larger than those of FLEXFIS and AMRules. Table \ref{tbl.Comparing.TimeAllMethodsAllDataSetsMilliseconds} shows that TSK-Streams is also a bit slower on average. We would argue, however, that this is not important, as it is still extremely fast in terms of absolute runtime: Being able to predict and learn from each new instance in just a few milliseconds, is certainly meets the requirements for learning on data streams.

	\begin{table*}[htp]
	\scriptsize
	\centering
	\caption{Performance of the different TSK-Streams variants in terms of RMSE.}
	\label{tbl.Comparing.performanceTSK-Streamsvariants}
	\begin{center}
		\resizebox{\linewidth}{!}{%
			\begin{tabular}{ p{0.12\linewidth}  p{0.19\linewidth} p{0.03\linewidth} | p{0.19\linewidth} p{0.03\linewidth} | p{0.19\linewidth} p{0.03\linewidth} | p{0.19\linewidth} p{0.03\linewidth}}	
				\hline
				Dataset&TSK-Streams&\multirow{2}{*}{\rotatebox[origin=c]{90}{Rank}}&TSK-Streams&\multirow{2}{*}{\rotatebox[origin=c]{90}{Rank}}&TSK-Streams&\multirow{2}{*}{\rotatebox[origin=c]{90}{Rank}}&TSK-Streams&\multirow{2}{*}{\rotatebox[origin=c]{90}{Rank}}\\
				&Error Red&&Error Red.&&Variance Red.&&Variance Red.&\\
				&One Cand.&&Both Cand.&&One Cand.&&Both Cand.&\\
				\hline
				2dplanes&$1.181 (0.031)$&$3$&$1.020 (0.001)$&$2$&$1.192 (0.042)$&$4$&$1.019  (0.003)$&$1$\\
				ailerons&$1.968 \stimes 10^{-4}(10^{-6})$&$1$&$2.594\stimes 10^{-4} (10^{-5})$&$2$&$1.351\stimes 10^{-3} (10^{-3})$&$4$&$4.166\stimes 10^{-4}(10^{-4})$&$3$\\
				bank8FM&$3.758\stimes 10^{-2} ( 10^{-4})$&$2$&$3.849\stimes 10^{-2} (10^{-4})$&$4$&$3.805\stimes 10^{-2} (10^{-3})$&$3$&$3.468\stimes 10^{-2}  (10^{-4})$&$1$\\
				calhousing&$70359 (334)$&$2$&$85976 (7054)$&$4$&$70639 (262)$&$3$&$69025 (190)$&$1$\\
				elevators&$0.003  (5.8 \stimes 10^{-5})$&$1$&$0.003 (3.7 \stimes 10^{-5})$&$2$&$0.005 (5.9 \stimes 10^{-4})$&$3$&$0.007 (9.9 \stimes 10^{-4})$&$4$\\
				fried&$2.283 (0.057)$&$3$&$2.253 (0.054)$&$2$&$2.448 (0.019)$&$4$&$2.220  (0.009)$&$1$\\
				house16h&$45927 (743)$&$3$&$45046 (579)$&$2$&$50107 (1968)$&$4$&$44721  (771)$&$1$\\
				house8&$41110 (1147)$&$1$&$45190 (2448)$&$2$&$91106 (21705)$&$4$&$69907 (8946)$&$3$\\
				kin8nm&$0.196  (10^{-3})$&$1$&$0.205(10^{-3})$&$4$&$0.203 (10^{-4})$&$3$&$0.201 (10^{-4})$&$2$\\
				mvnumeric&$1.779 (0.090)$&$3$&$0.741  (0.052)$&$1$&$2.347 (0.199)$&$4$&$1.092 (0.073)$&$2$\\
				pol&$31.276 (0.632)$&$4$&$24.297 (0.903)$&$2$&$24.813 (0.829)$&$3$&$17.733  (0.136)$&$1$\\
				puma32H&$0.0271 (1\stimes 10^{-4})$&$3$&$0.0274 (2.5\stimes 10^{-5})$&$4$&$0.025 (5\stimes 10^{-4})$&$2$&$0.019  (6.9\stimes 10^{-4})$&$1$\\
				puma8NH&$4.460 (0.036)$&$3$&$4.561 (0.003)$&$4$&$4.248 (0.029)$&$2$&$3.748  (0.040)$&$1$\\
				ratingssw.&$1.622 (0.001)$&$4$&$1.619 (0.001)$&$2$&$1.619 (0.001)$&$3$&$1.619 (0.001)$&$1$\\
				BNG-stock&$4.057 (0.028)$&$2$&$4.262 (0.036)$&$4$&$4.076 (0.024)$&$3$&$3.845 (0.014)$&$1$\\
				BNG-chol.&$49.327 (0.039)$&$3$&$49.582 (0.033)$&$4$&$49.310 (0.016)$&$2$&$49.116  (0.020)$&$1$\\
				BNG-echoM&$11.895 (0.021)$&$2$&$12.009 (0.009)$&$4$&$11.926 (0.007)$&$3$&$11.849 (0.009)$&$1$\\
				BNG-wine.&$0.780 (6.4 \stimes 10^{-4})$&$2$&$0.789 (10^{-4})$&$4$&$0.784 (3.9\stimes 10^{-4})$&$3$&$0.779  (1.7 \stimes 10^{-4})$&$1$\\
				\hline
				$\varnothing$ \textbf{Rank}&&$2.47$&&$2.97$&&$2.94$&&$1.61$\\
				\hline 
				\textbf{Wins/Losses}&$4/9$&&$3/9$&&$1/14$&&N/A \\
				\hline
			\end{tabular}
		}
	\end{center}
\end{table*}

\begin{table*}[htp]
	\scriptsize
	\centering
	\caption{Performance comparison of the algorithms with TSK-Streams (variance reduction with the extension for both candidates), in terms of RMSE.}
	\label{tbl.Comparing.performanceVsVarianceReductionBothCand}
	\begin{center}
		\resizebox{\linewidth}{!}{%
			\begin{tabular}{p{0.12\linewidth}  p{0.14\linewidth} p{0.03\linewidth} | p{0.14\linewidth} p{0.03\linewidth} |
				p{0.14\linewidth} p{0.03\linewidth} |  p{0.14\linewidth} p{0.03\linewidth} | p{0.14\linewidth} p{0.03\linewidth}}	
				\hline
				Dataset&AMRules&\rotatebox[origin=c]{90}{Rank}&FIMTDD&\rotatebox[origin=c]{90}{Rank}&ARF-Reg&\rotatebox[origin=c]{90}{Rank}&FLEXFIS&\rotatebox[origin=c]{90}{Rank}&TSK-Streams&\rotatebox[origin=c]{90}{Rank}\\
				\hline
				2dplanes&$1.385 (10^{-2})$&$2$&$2.621 (0.034)$&$4$&$3.65 (0.25)$&$5$&$2.389 (3.7\stimes 10^{-4})$&$3$&$1.019  (0.003)$&$1$\\
				ailerons&$5.1\stimes 10^{-4} (10^{-5})$&$4$&$7.9\stimes 10^{-4} (10^{-4})$&$5$&$3.8\stimes 10^{-4} (10^{-5})$&$2$&$1.9\stimes 10^{-4}  (10^{-7})$&$1$&$4.2\stimes 10^{-4} (10^{-4})$&$3$\\
				bank8FM&$3.6\stimes 10^{-2} (10^{-4})$&$2$&$1.2\stimes 10^{-1} (10^{-3})$&$5$&$1.1\stimes 10^{-1} (0.01)$&$4$&$3.7\stimes 10^{-2} (10^{-4})$&$3$&$3.5\stimes 10^{-2}  (10^{-4})$&$1$\\
				calhousing&$68336 (158)$&$2$&$82038 (225)$&$4$&$84179 (2043)$&$5$&$67177  (256)$&$1$&$69025 (190)$&$3$\\
				elevators&$0.008 (1 \stimes 10^{-3})$&$5$&$0.008 (6\stimes 10^{-4})$&$4$&$0.007 (4 \stimes 10^{-4})$&$3$&$0.004 (10^{-5})$&$1$&$0.007 (0.001)$&$2$\\
				fried&$2.404 (7.7\stimes 10^{-3})$&$2$&$3.490 (0.08)$&$4$&$3.88 (0.16)$&$5$&$2.635 (3\stimes 10^{-4})$&$3$&$2.220  (0.009)$&$1$\\
				house16h&$45621 (274)$&$2$&$50158 (383)$&$5$&$47883  (464)$&$3$&$48401 (67)$&$4$&$44721  (771)$&$1$\\
				house8&$40866 (564)$&$2$&$44613 (525)$&$4$&$41331  (1044)$&$3$&$40355  (894)$&$1$&$69907 (8946)$&$5$\\
				kin8nm&$0.204 (2\stimes 10^{-4})$&$3$&$0.272 (2\stimes 10^{-3})$&$5$&$0.23  (0.006)$&$4$&$0.203 (1\stimes 10^{-4})$&$2$&$0.201 (2\stimes 10^{-4})$&$1$\\
				mvnumeric&$2.689 (0.074)$&$2$&$4.768 (0.049)$&$4$&$8.19 (0.45)$&$5$&$3.349 (0.2)$&$3$&$1.092 (0.073)$&$1$\\
				pol&$19.956 (0.725)$&$2$&$29.173 (0.547)$&$3$&$42.51  (0.46)$&$4$&$59.028 (0.819)$&$5$&$17.733  (0.136)$&$1$\\
				puma32H&$0.021 (4.5 \stimes 10^{-4})$&$2$&$0.044 (3.3 \stimes 10^{-4})$&$5$&$0.037 (3\stimes 10^{-4})$&$4$&$0.030 (4.7\stimes 10^{-5})$&$3$&$0.019  (0.001)$&$1$\\
				puma8NH&$4.090 (0.032)$&$2$&$6.031 (0.034)$&$5$&$4.24  (0.19)$&$3$&$4.469 (0.004)$&$4$&$3.748  (0.04)$&$1$\\
				ratingssw.&$1.544 (3.1\stimes 10^{-3})$&$3$&$1.522  (7.5\stimes 10^{-3})$&$2$&$1.49  (0.005)$&$1$&$1.607 (1.3\stimes 10^{-3})$&$4$&$1.619 (0.001)$&$5$\\
				BNG-stock&$3.762  (0.011)$&$1$&$4.985 (0.035)$&$5$&$4.66  (0.11)$&$4$&$3.870 (0.044)$&$3$&$3.845 (0.014)$&$2$\\
				BNG-chol.&$49.727 (0.007)$&$3$&$50.516 (0.029)$&$4$&$51.35  (0.072)$&$5$&$49.537 ( 10^{-3})$&$2$&$49.116  (0.020)$&$1$\\
				BNG-echoM&$12.203 (0.022)$&$3$&$17.396 (0.09)$&$5$&$15.74  (0.08)$&$4$&$11.815  (6\stimes 10^{-3})$&$1$&$11.849 (0.009)$&$2$\\
				BNG-wine.&$0.786 (4 \stimes 10^{-4})$&$2$&$0.829 (3\stimes 10^{-3})$&$5$&$0.828  (0.007)$&$4$&$0.788 ( 10^{-4})$&$3$&$0.779  (2\stimes 10^{-4})$&$1$\\
				\hline
				$\varnothing$ \textbf{Rank}&&$2.44$&&$4.33$&&$3.77$&&$2.62$&&$1.83$ \\
				\hline
				\textbf{Wins/Losses}&$3/10$&&$2/15$&&$2/14$&&$6/11$&&N/A& \\
				\hline
			\end{tabular}
		}
	\end{center}
\end{table*}

	\begin{table*}[htp]
	\scriptsize
	\centering
	\caption{Size of the learned model of the algorithms in comparison with TSK-Streams (variance reduction with the extension for both candidates).}
	\label{tbl.Comparing.SizeVsVarianceReductionBothCand}
	\begin{center}
		\resizebox{1\linewidth}{!}{%
			\begin{tabular}{ p{0.12\linewidth} p{0.14\linewidth} p{0.03\linewidth} | p{0.14\linewidth} p{0.03\linewidth} | p{0.14\linewidth} p{0.03\linewidth} | p{0.14\linewidth} p{0.03\linewidth} | p{0.14\linewidth} p{0.03\linewidth}}
				\hline
				&AMRules&\rotatebox[origin=c]{90}{Rank}&FFIMTDD&\rotatebox[origin=c]{90}{Rank}&ARF-Reg&\rotatebox[origin=c]{90}{Rank}&FLEXFIS&\rotatebox[origin=c]{90}{Rank}&TSK-Streams&\rotatebox[origin=c]{90}{Rank}\\
				\hline
				2dplanes&$31.9(0.699)$&$2$&$140.9(0.830)$&$4$&$4084(4)$&$5$&$1(0)$&$1$&$36.4(2.499)$&$3$\\
				ailerons&$5.4(0.155)$&$2$&$24(0.894)$&$4$&$2304(100)$&$5$&$1(0)$&$1$&$8.6(0.253)$&$3$\\
				bank8FM&$8.1(0.386)$&$2$&$26.8(0.580)$&$4$&$1636(5)$&$5$&$1(0)$&$1$&$16.3(0.567)$&$3$\\
				calhousing&$10(0.245)$&$3$&$64.5(0.962)$&$4$&$4090(16)$&$5$&$1.7(0.145)$&$1$&$9.4(0.322)$&$2$\\
				elevators&$5(0.141)$&$2$&$40.4(1.401)$&$4$&$3197(81)$&$5$&$1(0)$&$1$&$9.5(0.604)$&$3$\\
				fried&$18.1(0.457)$&$3$&$119.6(1.644)$&$4$&$8076(11)$&$5$&$1.6(0.290)$&$1$&$12.1(0.411)$&$2$\\
				house16h&$6.1(0.095)$&$2$&$64.9(1.424)$&$4$&$4465(22)$&$5$&$1(0)$&$1$&$10(0.283)$&$3$\\
				house8&$6.4(0.210)$&$2$&$71.8(1.047)$&$4$&$4540(19)$&$5$&$3.1(0.754)$&$1$&$9.6(0.290)$&$3$\\
				kin8nm&$4.9(0.095)$&$2.5$&$24(0.787)$&$4$&$1635.9(3)$&$5$&$1.6(0.253)$&$1$&$4.9(0.170)$&$2.5$\\
				mvnumeric&$24.5(0.620)$&$2$&$130.2(1.223)$&$4$&$8170(189)$&$5$&$3.6(0.533)$&$1$&$33.2(3.786)$&$3$\\
				pol&$7.7(0.318)$&$2$&$43(1.020)$&$4$&$2453(90)$&$5$&$1(0)$&$1$&$26.3(1.504)$&$3$\\
				puma32H&$11.1(0.435)$&$2$&$28.3(0.401)$&$4$&$1559(3)$&$5$&$1(0)$&$1$&$22.4(1.409)$&$3$\\
				puma8NH&$6.7(0.348)$&$2$&$28.1(0.624)$&$4$&$1643(5)$&$5$&$1.1(0.095)$&$1$&$10.5(0.570)$&$3$\\
				ratingssw.&$9.9(0.221)$&$3$&$63.2(1.121)$&$4$&$3775(6)$&$5$&$7.8(4.711)$&$2$&$2(0)$&$1$\\
				BNG-stock&$16.6(0.210)$&$3$&$178.6(2.164)$&$4$&$11726(9)$&$5$&$6.5(1.790)$&$1$&$13.3(0.511)$&$2$\\
				BNG-chol.&$129.7(0.633)$&$3$&$3150.1(6.981)$&$4$&$209657(1535)$&$5$&$1(0)$&$1$&$42(0.346)$&$2$\\
				BNG-echoM&$8.5(0.255)$&$2$&$55.3(0.801)$&$4$&$3633(30)$&$5$&$1.5(0.474)$&$1$&$9.5(1.492)$&$3$\\
				BNG-wine.&$74.8(0.597)$&$3$&$1533.9(3.659)$&$4$&$104532(36)$&$5$&$1.1(0.095)$&$1$&$28.7(0.247)$&$2$\\
				\hline
				$\varnothing$ \textbf{Rank}&&$2.36$&&$4.0$&&$5.0$&&$1.05$&&$2.58$\\				
				\hline 
			\end{tabular}
		}
	\end{center}
\end{table*}

	\begin{table*}[htp]
		\scriptsize
		\centering
		\caption{Average milliseconds needed to process each instance (training + testing).}
		\label{tbl.Comparing.TimeAllMethodsAllDataSetsMilliseconds}
		\begin{center}
			\resizebox{1\linewidth}{!}{%
				\begin{tabular}{ p{0.12\linewidth} p{0.2\linewidth}  p{0.2\linewidth} p{0.2\linewidth}  p{0.2\linewidth}  p{0.2\linewidth}  }     
					\hline
					&AMRules&FIMTDD&ARF-Reg&FLEXFIS&TSK-Streams\\
					\hline
					2dplanes&$0.192 (0.006)$&$0.033 (0.001)$&$0.079 (0.002)$&$1.329 (0.004)$&$4.478 (0.297)$\\
					ailerons&$0.260 (0.010)$&$0.133 (0.007)$&$0.252 (0.003)$&$2.118 (0.004)$&$4.752 (0.201)$\\
					bank8FM&$0.201 (0.009)$&$0.128 (0.004)$&$0.097 (0.001)$&$1.812 (0.025)$&$1.874 (0.096)$\\
					calhousing&$0.179 (0.006)$&$0.267 (0.011)$&$0.102 (0.001)$&$1.503 (0.044)$&$1.205 (0.052)$\\
					elevators&$0.189 (0.007)$&$0.109 (0.006)$&$0.138 (0.001)$&$1.517 (0.002)$&$2.059 (0.117)$\\
					fried&$0.240 (0.011)$&$0.791 (0.035)$&$0.112 (0.001)$&$1.389 (0.039)$&$1.900 (0.143)$\\
					house16h&$0.185 (0.006)$&$0.580 (0.037)$&$0.144 (0.001)$&$1.470 (0.037)$&$2.199 (0.108)$\\
					house8&$0.155 (0.005)$&$0.272 (0.011)$&$0.099 (0.001)$&$1.774 (0.116)$&$1.093 (0.024)$\\
					kin8nm&$0.199 (0.006)$&$0.149 (0.008)$&$0.101 (0.001)$&$1.993 (0.052)$&$0.632 (0.031)$\\
					mvnumeric&$0.248 (0.009)$&$0.604 (0.030)$&$0.103 (0.002)$&$2.193 (0.163)$&$3.481 (0.255)$\\
					pol&$0.267 (0.013)$&$0.135 (0.006)$&$0.307 (0.001)$&$2.531 (0.030)$&$9.799 (0.674)$\\
					puma32H&$0.426 (0.022)$&$0.623 (0.028)$&$0.247 (0.001)$&$2.478 (0.010)$&$8.539 (0.696)$\\
					puma8NH&$0.207 (0.010)$&$0.137 (0.005)$&$0.105 (0.001)$&$2.630 (0.439)$&$1.148 (0.091)$\\
					ratingssw.&$0.145 (0.006)$&$0.031 (0.002)$&$0.057 (0.001)$&$2.561 (0.648)$&$0.086 (0.002)$\\
					BNG-stock&$0.235 (0.007)$&$1.190 (0.046)$&$0.111 (0.001)$&$1.632 (0.163)$&$1.660 (0.056)$\\
					BNG-chol.&$0.864 (0.039)$&$10.047 (0.373)$&$0.180 (0.006)$&$1.053 (0.002)$&$7.860 (0.230)$\\
					BNG-echoM&$0.164 (0.005)$&$0.199 (0.011)$&$0.121 (0.007)$&$1.413 (0.049)$&$1.333 (0.257)$\\
					BNG-wine.&$0.710 (0.040)$&$15.351 (0.675)$&$0.169 (0.001)$&$1.044 (0.013)$&$5.043 (0.236)$\\
					\hline
				\end{tabular}
			}
		\end{center}
	\end{table*}

	\section{Conclusion}
	
	In this paper, we introduced a new fuzzy rule learner for adaptive regression on data streams, called TSK-Streams. This method combines the effectivity of concepts for rule induction as implemented in AMRules with the expressivity of TSK fuzzy rules. TSK-Streams as presented in this paper is an improved variant of an earlier version \citep{mpub360}; modifications essentially concern all parts of the learning algorithm, including the discretization, the rule extension, and the drift detection. 
	
	In an experimental study with real and synthetic data, we compared TSK-Streams with state-of-the-art regression algorithms for learning from data streams: AMRules, FIMTDD, ARF-Reg and FLEXFIS. The results are very promising, especially because our learner achieves the best performance in terms of predictive accuracy. This is remarkable, given that AMRules and FLEXFIS are truly strong (and indeed still competitive) learners---these methods have been developed over many years, and are therefore difficult to beat.


	Our current implementation of TSK-Streams can be obtained from our Github repository\footnote{\url{https://github.com/shaker82/TSK-Streams}}.


\end{document}